\documentclass[10pt,twocolumn,letterpaper]{article}

\usepackage[pagenumbers]{cvpr} 

\usepackage{graphicx}
\usepackage{amsmath}
\usepackage{amssymb}
\usepackage{booktabs}

\usepackage[pagebackref=true,breaklinks=true,colorlinks=true,bookmarks=false,linkcolor={red!50!black},urlcolor={darkgray!80!black},citecolor={green!50!black}]{hyperref} 

\usepackage[capitalize]{cleveref}
\crefname{section}{Sec.}{Secs.}
\Crefname{section}{Section}{Sections}
\Crefname{table}{Table}{Tables}
\crefname{table}{Tab.}{Tabs.}

\usepackage{xcolor}
\usepackage{enumitem}
\usepackage{xspace}
\usepackage{booktabs}
\usepackage{colortbl}
\usepackage{multirow}
\usepackage{makecell}
\usepackage{lipsum} 
\usepackage{gensymb} 

\usepackage[labelsep=period]{caption}
\renewcommand{\paragraph}[1]{\vspace{0.2em}\noindent \textbf{#1 \hspace{0.2em}}}

\definecolor{MyDarkRed}{rgb}{0.46, 0.16, 0.16}
\definecolor{MyDarkBlue}{rgb}{0.16, 0.16, 0.66}

\def\confName{Technical Report}

\begin{document}

\title{GauStudio: A Modular Framework for 3D Gaussian Splatting and Beyond}

\author{Chongjie Ye\textsuperscript{\rm 1,2}$^{*}$, Yinyu Nie\textsuperscript{\rm 3}$^{*}$, Jiahao Chang\textsuperscript{\rm 1, 2}$^{*}$, Yuantao Chen\textsuperscript{\rm 2}, Yihao Zhi\textsuperscript{\rm 1,2}, Xiaoguang Han\textsuperscript{\rm 2,1}$^{\dag}$ \\
\small{$^{*}$equal contribution} \qquad \small{$^{\dag}$corresponding author} \vspace{5pt}\\
\textsuperscript{\rm 1}{FNii, CUHKSZ} \qquad \textsuperscript{\rm 2}{SSE, CUHKSZ} \qquad \textsuperscript{\rm 3}{Technical University of Munich} \qquad 
\vspace{5pt}\\
\small{\href{https://github.com/GAP-LAB-CUHK-SZ/gaustudio}{github.com/GAP-LAB-CUHK-SZ/gaustudio}}
}




\twocolumn[{%
\renewcommand\twocolumn[1][]{#1}%
\maketitle
\begin{center}
  {
  \captionsetup{type=figure}
  \includegraphics[width=1.\linewidth]{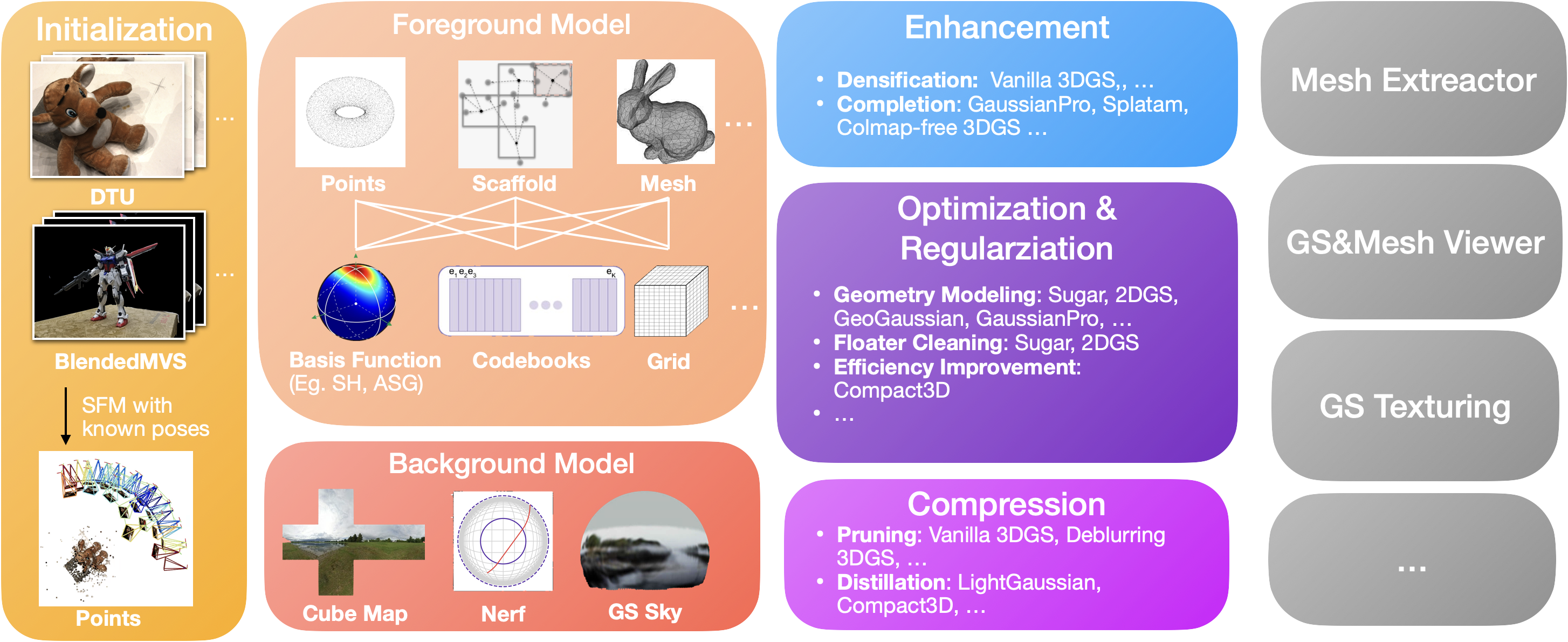}
  \captionof{figure}{
  \textbf{GauStudio} is a modular framework that unifies various 3D Gaussian Splatting techniques. It decomposes 3D scenes into components like foreground and background models, represented using specialized techniques on 3D Gaussians. These components can be flexibly combined and rendered to synthesize novel views, enabling tailored modeling pipelines for different tasks.
  }
  \label{fig:teaser}
}
\end{center}
}]


\begin{abstract}
    We present GauStudio, a novel modular framework for modeling 3D Gaussian Splatting (3DGS) to provide standardized, plug-and-play components for users to easily customize and implement a 3DGS pipeline. Supported by our framework, we propose a hybrid Gaussian representation with foreground and skyball background models. Experiments demonstrate this representation reduces artifacts in unbounded outdoor scenes and improves novel view synthesis. Finally, we propose Gaussian Splatting Surface Reconstruction (GauS), a novel render-then-fuse approach for high-fidelity mesh reconstruction from 3DGS inputs without fine-tuning. Overall, our GauStudio framework, hybrid representation, and GauS approach enhance 3DGS modeling and rendering capabilities, enabling higher-quality novel view synthesis and surface reconstruction.
\end{abstract}

\section{Introduction}
\label{sec:intro}
Novel view synthesis - the task of rendering photorealistic novel views of a 3D scene from partial image observations - is a longstanding challenge in computer vision and graphics. Recent advances in neural rendering techniques like NeRF~\cite{blender} have made significant strides by optimizing multi-layer perceptrons (MLPs) using volumetric ray-marching to represent and synthesize novel views of captured scenes. Recently, many efficient radiance field follow-ups are built on continuous representations by interpolating values stored in voxel grids~\cite{fridovich2022plenoxels, wu2022voxurf}, hash tables~\cite{muller2022instant}, or point-based data structures~\cite{xu2022point,sandstrom2023point}.

However, these implicit neural representations often struggle to model high-frequency details and lack explicit geometric structures, making them difficult to edit and compress for real-time applications. As an alternative, explicit 3D representations such as 3D Gaussians\cite{3dgs} have recently emerged as an efficient and editable scene modeling approach. Unlike NeRFs\cite{mildenhall2021nerf,mip-nerf360} which model a continuous volumetric field, 3D Gaussians Splatting(3DGS) represents a scene as a set of anisotropic 3D Gaussian kernels which encode local geometry and view-dependent appearance. During testing time, these Gaussians can reproduce detailed geometry and view effects with a rendering and editing friendly format.

In this work, we introduce GauStudio, a modular framework tailored for 3D Gaussian Splatting (3DGS) with a set of customizable modules for diverse scene reconstruction scenarios. The key advantage of GauStudio lies in its highly modularized design. Its architecture allows different components to be flexibly combined and replaced, enabling the convenient construction of different 3D scene modeling methods. Researchers can freely combine different foreground models (e.g., Gaussian models), background models (e.g., environment maps~\cite{2311.17977v1} or our Gaussian Sky models), and other components according to their needs, forming hybrid modeling pipelines tailored for specific scenes and tasks. This composability significantly enhances the framework's flexibility and scalability, facilitating accelerated innovation in the field of 3D scene modeling.

GauStudio consists of several key stages: 1) scene initialization from SFM~\cite{schonberger2016structure} or from splatting priors~\cite{2312.12337v2}; 2) Gaussians' optimization with geometric and sparsity regularizers; 3) an enhancement step to boost 3D Gaussians' representation ability; 4) scene compression via learnable or geometric pruning. We also consider different representations for encoding view-dependent appearance within the Gaussian splats, including the traditional spherical harmonics, learned neural feature vectors, and explicit feature caching structures. 

Additionally, we present an efficient surface reconstruction module from optimized 3D Gaussians, namely \textit{GauS}. It uses volumetric fusion to convert 3D Gaussians into textured triangle meshes. Combined with our GauStudio which supports various GS-based methods~\cite{2402.14650v1, 2311.16493v1, lu2023scaffold, 2312.09147v2} natively, our Gaus is a plug-and-play module that can easily extract meshes for different types of Gaussians with a single command. For background modeling, we build the sky background using spherical environment maps composed of Gaussians. All these modules are fully customizable in GauStudio.

In summary, we introduce GauStudio, a flexible and modular framework assembled with 3D Gaussian Splatting techniques for different tasks (e.g., scene reconstruction, editing, simulation, and path planning). The key contributions are concluded as follows:

\begin{enumerate}
   \item A modularized and composable Gaussian Splatting framework for practitioners to seamlessly integrate different components, e.g., foreground Gaussian models, background representations, and other modules based on their specific tasks.
   
   \item The framework encapsulates many key stages in 3D Gaussians reconstruction, such as initialization, optimization, regularization, representation enhancement, and compression, with each stage fully customizable. 
   
   \item We provide an efficient mesh extraction pipeline, namely \textit{GauS}, to convert optimized Gaussians into textured meshes, serving as a versatile plug-and-play solution across different GS-based methods.
\end{enumerate}

Overall, GauStudio presents a comprehensive yet customizable platform to drive innovation in 3D scene modeling tasks like reconstruction, editing, and simulation.



\section{Preliminaries}
\label{sub:Preliminaries}
\textbf{Neural volume rendering (NeRF)} NeRF~\cite{mildenhall2021nerf} represents a scene through continuous volumetric density and color fields. To render a view, NeRF first casts the rays from a posed camera and samples 3D points along the rays. Then, it predicts the volume density $\sigma$ and RGB color $c$ at 3D positions $x$ by querying the multi-layer perceptrons (MLPs), and the color is integrated along each ray. Formally, The rendered color $\hat{\mathbf{c}}(\mathbf{o},\mathbf{d})$ for a pixel approximates a Riemann sum over N samples: 
\begin{equation}
\hat{\mathbf{c}}(\mathbf{o},\mathbf{d}) = \sum_{i=1}^N w_i \mathbf{c}i,
\end{equation}
where the weight $w_i$ is determined by the accumulated opacity along each ray. The color loss $\mathcal{L}_\text{RGB}$ which measures the difference between renderings and input views provides supervision for the MLPs predicting $\sigma$ and $\mathbf{c}$:
\begin{equation}
\mathcal{L}_\text{RGB} = | \hat{\mathbf{c}} - \mathbf{c} |_1.
\end{equation}
\label{equ:RGBLoss}

\textbf{3D Gaussian Splatting (3DGS)} Recent works have developed Gaussians as an explicit 3D scene representation for novel view synthesis~\cite{3dgs,2311.16493v1,2402.14650v1,lu2023scaffold}. In contrast to implicit NeRFs~\cite{mildenhall2021nerf}, 3DGS represents a scene using a set of 3D anisotropic Gaussian kernels. Each kernel is defined by: Mean position $\mu$, Opacity $\sigma$, Covariance matrix $\sum$, and Spherical harmonic (SH) coefficients $C$ for representing angular color variations. To render a view, the Gaussian splats are first projected onto 2D screen space based on sorted depth. Then, differentiable volumetric rendering combines the kernels' color and density properties, which can be formulated as:
\begin{equation}
C = \sum_{k\in\mathcal{P}}\alpha_{k}\text{SH}(d_{k}; \mathbf{C}{k}) \prod_{j=1}^{k-1}(1 - \alpha_{j})
\end{equation}

\section{GauStudio}
\label{sec:method}
In this section, we describe the components of GauStudio and how they are integrated into a uniform pipeline. An overview of our framework is exhibited in Fig.\ref{fig:teaser}. In particular, our framework enables a hybrid scene representation comprising Gaussian foreground models and background models. As for the foreground Gaussians, GauStudio can be readily converted them to textured meshes for explicit surface modeling, obviating the need for further fine-tuning or NeRF-based~\cite{mildenhall2021nerf}  re-training. Meanwhile, the background models support diverse backends like environment maps, textured meshes, and other neural representations~\cite{muller2022instant}. 

\subsection{Problem setting} 
Given a set of calibrated RGB images $I = \{I_1, \ldots, I_N\}$ and the corresponding camera parameters $\Theta = \{\theta_1, \ldots, \theta_N\}$, where $\theta_i = \{R_i, T_i, K_i\}$ represents the extrinsic rotation $R_i \in \mathbb{R}^{3 \times 3}$ (an element of the Special Orthogonal group SO(3)), extrinsic translation $T_i \in \mathbb{R}^3$, and intrinsic camera parameters $K_i = [c_x, c_y, f_x, f_y]^\top$ with principal point $(c_x, c_y)$ and focal lengths $(f_x, f_y)$, GauStudio aims to learn an optimized  radiance field $F^*$ that models the scene's appearance and geometry.

We define the 3D Gaussian Splatting (3DGS) function $F$ as a mapping from a 3D spatial position $\mathbf{x} \in \mathbb{R}^{3}$, view direction $\mathbf{\theta}$ and $\mathbf{\phi}$, and camera parameters $\Theta = \{\mathbf{R}, \mathbf{T}, \mathbf{K}\} \in \mathbb{R}^{3} \times \mathbb{R}^{3} \times \mathbb{R}^{4}$ to a color $\mathbf{c} \in \mathbb{R}^{3}$ and density $\sigma \in \mathbb{R}$:

$$F: \mathbb{R}^{3} \times \mathbb{R}^{2} \times (\mathbb{R}^{3} \times \mathbb{R}^{3} \times \mathbb{R}^{4}) \rightarrow \mathbb{R}^{3} \times \mathbb{R}$$
$$F(\mathbf{x}, \mathbf{d}, \Theta) = (\mathbf{c}, \sigma)$$

The goal is to find the optimal radiance field $F^*$ that minimizes the image reconstruction loss and regularization terms:

$$F^* = \operatorname{argmin}{F} \mathcal{L}_{\text{rgb}}(F; I, \Theta) + \lambda \mathcal{L}_{\text{reg}}(F)$$

Where $\mathcal{L}_{\text{rgb}}$ is the image reconstruction loss, and $\mathcal{L}_{\text{reg}}$ represents additional regularization losses. We postpone the details of $\mathcal{L}_{\text{rgb}}(F; I, \Theta)$ and $\mathcal{L}_{\text{reg}}(F)$ to Section \ref{sub:regularization}.



\subsection{Framework overview}
GauStudio implements 3D Gaussian Splatting (3DGS) through a unified reconstruction framework $\varphi$ that takes input images $I$ and camera projection matrices $P$ calculated from camera parameters $\Theta = {\theta_1, \ldots, \theta_N}$ as input and outputs optimized 3D Gaussian Kernels $X$ and their parameters $\gamma$:
\begin{align}
\varphi(I, P) = X, \theta \\
F(\mathbf{x}, \mathbf{d}, \Theta; X, \gamma) = (\mathbf{c}, \sigma)
\end{align}

Our framework $\varphi$ comprises four components: Initialization, Optimization, Enhancement and Compression.

\textbf{Initialization} The Initialization module $\psi$ takes the input image $I$, the projection function $P$, and an optional set of parameters $\Omega_\psi$ (e.g., initialization strategy, desired number of Gaussians) to generate the initial Gaussians $X$ and their parameters $\theta$:$$X^{(0)}, \gamma^{(0)} = \psi(I, P; \Omega_\psi)$$

\textbf{Optimization} The Optimization stage $\omega$ takes the initial Gaussians $X$, parameters $\theta$, input image $I$, projection function $P$, and an optional set of parameters $\Omega_\omega$ (e.g., regularization terms, depth priors) to produce the optimized Gaussians: $\hat{X}$ and parameters $\hat{\theta}$. $$ X^{(1)}, \gamma^{(1)} = \omega(X^{(0)}, \theta^{(0)}, I, P; \Omega_\omega)$$

\textbf{Enhancement} Unlike neural radiance fields, which implicitly capture the entire scene, the Gaussian Kernels representation requires the existence of corresponding Gaussian Kernels to accurately model specific parts of the scene. Additionally, gaussian's representation ability is bounded by the amounts of kernels and the method to model view-dependent appearance.  By densifying and completing the optimized Gaussians $\hat{X}$ and parameters $\hat{\theta}$, the Enhancement stage augments the representation with additional Gaussian Kernels, enabling it to model high-frequency details and provide a more complete representation of the scene: $$X^{(2)}, \gamma^{(2)} = \delta(X^{(1)}, \theta^{(1)}, P; \Omega_\delta)$$

\textbf{Compression} However, introducing a large number of Gaussian Kernels can lead to computational inefficiency and increased memory requirements, which can be detrimental to the overall performance of the system. To address this concern, the Compression stage becomes essential. The Compression stage aims to strike a balance between the representation's ability to capture scene details and its efficiency in terms of computational resources and memory usage. It achieves this by identifying and removing insignificant Gaussian Kernels that contribute minimally to the overall representation, or by compacting the parameters of the Gaussian Kernels through techniques such as quantization or pruning: $X^{(3)}, \gamma^{(3)} = \pi(X^{(2)}, \theta^{(2)}, P; \Omega_\pi)$

\subsection{Representation}
To modeling the view-dependent appearance of each 3D Gaussian, we consider various representations in our GauStudio framework as illustrated in Fig.\ref{fig:teaser}. In particular, we consider Fixed Basis Function, Neural Feature Vectors, Triplane, Hash Grid, Codebook.

\textbf{Fixed Basis Functions:} Traditional Gaussian splattingç utilizes spherical harmonics (SH) to model the view-dependent color. Additionally, we consider anisotropic spherical Gaussians (ASG)~\cite{yang2024spec} for modeling specular components within scenes and Spherical Gaussians (SG)\cite{yariv2023bakedsdf} for balancing image quality and rendering speed. To achieve material and lighting decomposition, we further consider associating each point with additional PBR parameters\cite{2312.03704v1, 2311.16473v2,2311.17977v1, 2312.05133v1, 2311.16043v1} and indirect illumination\cite{2311.17977v1, 2312.05133v1} within our framework.

\textbf{Neural Feature Vectors:} To model the view-dependent color in real-time view synthesis systems, each point is associated with a compact Neural Feature Vector, which is decoded using a small MLP\cite{2401.00834v1, lu2023scaffold, 2312.04564v1, 2311.18482v1}.

\textbf{Triplane, Hash Grid:} Instead of indexing neural feature vectors with point cloud IDs, alternative approaches cache the neural features in explicit data structures, such as triplanes\cite{2312.09147v2, 2401.04099v1} and hash grids\cite{lee2023compact}. For any given position $x$, the corresponding feature vector can be queried from these explicit structures. In the triplane representation, the position is projected onto axis-aligned feature planes, and the final feature $f_t$ is obtained by concatenating three trilinear interpolated features. For hash grids, the feature is interpolated from a multi-resolution grid stored using instant-ngp\cite{muller2022instant}. To support unbounded scenes, we consider employing the contraction technique\cite{yariv2023bakedsdf} to compactly represent background features.

\textbf{Codebook Quantization} Motivated by the balance between compact storage and rendering efficiency, we also consider learning a discrete codebook of vector codes during training\cite{2311.18159v1, 2311.17245v4}. This is crucial as only a simple lookup into the codebook is required using the stored indices at rendering time. We seek to exploit lower bandwidth and computational costs compared to conventional feature fetching and processing.

\subsection{Initialization}
ciTraditional 3D Gaussian splatting pipelines \cite{3dgs} often rely on a sparse point cloud initialized from correspondence-rich image pairs. However, this approach can be sensitive to the quality of feature matches and may fail to reconstruct texture-less surfaces accurately. Our GauStudio framework, besides traditional initializations from SfM point clouds \cite{schonberger2016structure}, MVS point clouds \cite{2402.00763v1, 2401.06003v1, 2311.16043v1}, depth maps \cite{fu2023colmap}, and LiDAR\cite{2312.07920v2,2401.01339v1}, we consider alternative strategies that better leverage off-the-shelf
prior network.

\textbf{Semi-Dense Point Cloud Initialization} Inspired by recent advances in semi-dense feature matching \cite{sun2021loftr, edstedt2023dkm} and point tracking \cite{karaev2023cotracker, wang2023tracking}, we propose initializing the point cloud using dense feature matches instead of traditional detector-based matches. Specifically, we extract and aggregate dense feature correspondences using hierarchical localization frorm\cite{sarlin2019coarse}, then form the point cloud via multi-view 3D triangulation using the Direct Linear Transform (DLT) \cite{hartley2003multiple}. Semi-dense initialization strikes a balance between training time (which can be prohibitively high for dense MVS point initialization) and rendering quality on texture-less surfaces and unbouneded background.

\textbf{Generalized Gaussian Splatting Initialization} In scenarios with limited scene observations, the optimized gaussian shapes may deviate significantly from the actual surface geometry. This deviation leads to a decline in the rendering quality when viewed from a new viewpoint. To alleviate this issue, we consider initializing the Gaussians using pre-computed properties $\theta$ from a generalized Gaussian splatting model\cite{2312.12337v2,  2401.04099v1, 2312.09147v2, 2312.13150v1, tang2024lgm} for further optimization.

\subsection{Optimization and Regularization}\label{sub:regularization}
In the simplest case, the Gaussian kernels are optimized to minimize the image reconstruction loss (Eq.\ref{equ:RGBLoss}). Additionally, auxiliary regularization can guide the optimization towards desired properties. We classify the regularization based on the optimization target: Geometry Modeling, Floater Cleaning, and Efficiency Improvement.

\textbf{Geometry Modeling} Extracting a mesh from millions of tiny 3D Gaussians is challenging due to their arbitrary spatial distribution and anisotropic kernels. One approach is to represent surfaces with Gaussians is to flatten them into 3D "surfels" aligned with the surface. For instance, adopting scaling loss \cite{2312.00846v1} could minimize the smallest component of the scaling factor $s \in \mathbb{R}^3$ for each Gaussian towards zero. This regularization also aids in determining Gaussian normals by selecting the shortest scaled edge. To align Gaussians with the surface, we consider regularizing the rendered normals against planar constraints \cite{2402.14650v1, 2402.00763v1}, rendered depth buffers \cite{2311.16043v1, 2311.16473v2, 2311.17977v1}, or normal priors \cite{li2023spec, yu2022monosdf}.

\textbf{Floaters Cleaning} We employ the entropy loss\cite{sugar} to encourage the Gaussians' opacities to be either 0 or 1. Then, with the pruning mechanism pipeline, floaters will be automatically removed during training. However, we found that such regularization could lead to a trade-off situation, as it may reduce the number of remaining points, potentially compromising the rendering quality and negatively impacting transparent components.

\textbf{Efficiency Improvement} Since small Gaussians have a negligible contribution to the overall rendering quality, it is natural to consider removing less-contributing Gaussians for efficiency with little rendering quality loss. We consider the masking loss\cite{lee2023compact} to encourage Gaussian sparsity. Learnable masking is utilized to eliminate redundant Gaussians based on their volume (scales $s$) and transparency (opacities $o$). Binary masks $M$ generated from a mask parameter $m$ are applied to $s$ and $o$, effectively removing Gaussians with small volumes or low opacities.

\subsection{Enhancement}
Gaussians Enhancement is critical, especially given sparse point cloud initialization. We follow \cite{3dgs} and \cite{2312.00451v1} and apply rule-based adaptive control during optimization. To better model textureless surfaces and improve convergence speed, monocular depth from an RGB-D sensor or a prior network\cite{yang2024depth} could be taken into account for complete the gaussians during training\cite{fu2023colmap, keetha2023splatam}. Additionally, a novel completing method has been proposed to utilize propagated depth, which serves the same effort as depth-guided densification while in a self-supervised way. For scaffold-based representations\cite{lu2023scaffold, yang2024spec}, a different densification method is adopted to control the number of anchor points according to the accumulated gradients within the voxel instead of screen spaces.

Besides traditional splitting, copying, and inserting methods, we tend to explore point cloud completion and upsampling methods to empower Gaussian densification in the future \cite{2312.09147v2}.

\subsection{Compression}
We consider a simple opacity-based pruning strategy for general learning. Additionally, if a mask is provided, a visible mask will be calculated to prune the Gaussians projected onto the masked pixels. We could also use depth for prune the 3D gaussians in the free space or control the density of gaussians according to the distance\cite{2401.00834v1}. 



\begin{figure}[h]
    \centering
    \includegraphics[width=\linewidth]{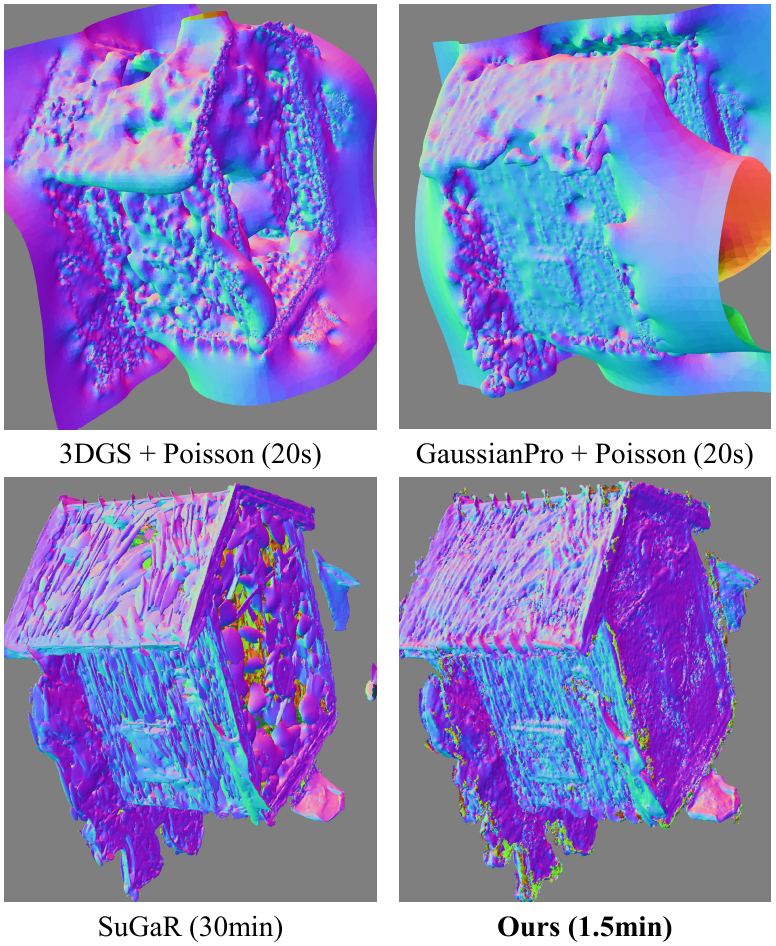}
    \caption{Comparsion with other mesh extraction strategies.
    Utilizing the Poisson reconstruction directly on 3DGS leads to a bad performance.
    Due to the special design of GaussianPro on normal and depth, the performance of Poisson reconstruction on GaussianPro is slightly better.
    SuGaR always tends to produce larger ellipses.
    Compared with the above methods, our method can generate high-quality mesh efficiently.}
    \label{fig:clock}
\end{figure}

\section{Gaussian Surface Reconstrution(GauS)}
Converting 3D Gaussians into triangle meshes presents a non-trivial task, and several strategies have been proposed to tackle this challenge. Among these, the most basic approach involves applying Poisson surface reconstruction \cite{poisson, screened} to the means of the 3D Gaussians, with point normals computed using the shortest axis of the Gaussians~\cite{sugar}. Nonetheless, this method often yields poor results due to the noisy Gaussians around the surface and leads to floater-like Gaussians in the free space.

Recently, Sugar \cite{sugar} proposed several geometry-oriented regularizations to align the Gaussians with the surface of the scene. Despite yielding promising results, this approach is plagued by long refinement periods because it relies on a computationally expensive sorting algorithm to determine the connectivity of neighboring Gaussians.

In our experiments with the Sugar \cite{sugar}, we noticed that even without fine-tuning and regularization, most points were well-aligned with the surface when dense training views are provided. However, due to noisy estimated normals, the resulting mesh quality suffered, displaying numerous holes and artifacts. While some of the regularization methods discussed in Sec.\ref{sub:regularization} could potentially enhance normal quality, they incur a 1-2 dB lower PSNR in object-centric and indoor scenes, representing an undesirable trade-off.

\begin{figure}[h]
    \centering
    \includegraphics[width=\linewidth]{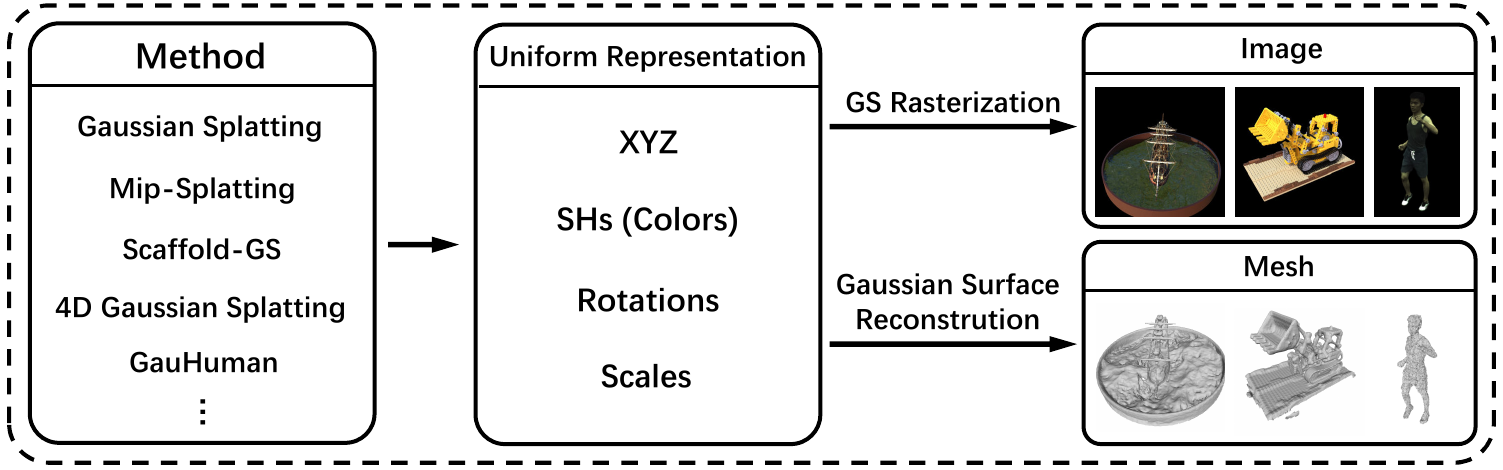}
    \caption{Our Gaussian Surface Reconstrution can be integrated into existing 3DGS pipelines.
    Before GS rasterization, 3DGS-based frameworks will convert Gaussian Splatting into a unified representation that can be used for our Gaussian Surface Reconstruction.
    }
    \label{fig:gaus}
\end{figure}


To circumvent the dependency on Gaussian normals and their associated challenges, we employed a volumetric fusion approach \cite{vdbfusion}. Specifically, we rendered the median depth as proposed in \cite{keetha2023splatam} and fused it into a mesh using the VDBFusion~\cite{vdbfusion}. Fig.\ref{fig:clock} demonstrates the effectiveness of volumetric fusion compared to naive Poisson reconstruction and the Sugar method \cite{sugar}, showcasing significantly improved quality and efficiency.

Notably, our surface reconstruction method is compatible with nearly all 3DGS-based methods, as illustrated in Fig.\ref{fig:gaus} by utilizing a customized GS rasterizer. This versatility ensures seamless integration with existing 3DGS pipelines and frameworks, facilitating widespread adoption and application.

Moreover, a compelling motivation for employing volumetric fusion lies in its scalability, memory efficiency, and time efficiency, particularly for super large-scale scenes like city districts and buildings. This attribute makes our approach ideally suited for handling intricate, large-scale environments, which is essential for numerous real-world applications across domains such as urban modeling, virtual reality, and digital twins.

\section{Gaussian Sky Modeling}
Outdoor scenes often contain sky regions located at an infinite distance from the camera. The 3DGS model tends to generate cloud-like artifacts in the foreground region when attempting to model the sky. These artifacts significantly influence the quality of novel view synthesis. To address this issue, our rendering model incorporates a spherical environment map composed of gaussians to model the sky separately from the foreground. We render the color for each pixel using compositional rendering similar to \cite{mars}:

\begin{align}
C(p) = \sum_i T_i\alpha_ic_i + (1 - O(p)) \cdot C_{sky}(p) \\
O(p) = \sum_i T_i\alpha_i \\
\quad T_i = \prod_{k=1}^{i-1}(1-\alpha_i)
\end{align}

Here, $C_{sky}(p)$ is the rendered pixel color from the sky model, $O(p)$ is the foreground opacity value obtained by summing the sample weights, and $T_i$ represents the accumulated transmittance.

However, we found that simply blending the sky color and foreground color could not guarantee the clearance of artifacts when sparse points existed in the free space initially. To address this, we incorporate semantic priors from a semantic segmentation network \cite{cheng2022masked} to generate sky masks. We then penalize the opacity values $O(p)$ for the sky region to be close to zero, enforcing a clear separation between the sky and foreground regions:

$$L_{sky} = \lambda_{sky} \sum_{p \in \mathcal{R}_{sky}} O(p)$$

where $\lambda_{sky}$ is the weighting factor. 

Experiments demonstrate that penalizing the mask loss in the sky region helps remove cloud artifacts, even with an imperfect sky mask. Although alternative approaches, such as a sky depth loss to penalize the sky region depth to be as far as possible or a binary cross-entropy (BCE) semantic regularization, were explored, we found that they tended to require longer convergence times and could adversely impact the foreground modeling quality.

Nonetheless, the sky depth loss and BCE semantic regularization could potentially be beneficial in specific scenarios or as complementary losses. The sky depth loss may aid in enhancing the perceived depth separation between the sky and foreground elements, while the BCE semantic regularization could further refine the sky-foreground boundary based on semantic cues.



\section{Experiments}
\label{sec:Experiments}
\subsection{Experimental Setup}
\textbf{Dataset and Metric} 
In order to assess the effectiveness of our gaussian-based environment map, we conducted an evaluation on commonly used Tanks and Temples~\cite{Knapitsch2017} dataset.
We did not perform experiments with Mip-NeRF360 \cite{mip-nerf360} and DeepBlending \cite{deepblending} since our method focuses on scenes with unbounded skies.
We quantified the image synthesis quality with standard metrics such as Peak Signal-to-Noise Ratio (PSNR), Structural Similarity Index Measure (SSIM)~\cite{ssim}, and Learned Perceptual Image Patch Similarity (LPIPS) \cite{lpips} to measure the effectiveness of the method.
Additionally, we also reported the memory footprint and runtime for each method to assess their computational efficiency.

In order to evaluate our gaussian surface reconstruction, we conducted expriments on the DTU~\cite{dtu} and Blender~\cite{blender} datasets. 
For the assessment of surface reconstruction, we present lots of qualitative comparisons.

\textbf{Baselines and Implementation} 
For our gaussian-based environment map, we selected 3D-GS~\cite{3dgs} as our main baseline and also included some latest 3DGS-based works, including Scaffold-GS \cite{lu2023scaffold}, GaussianPro \cite{2402.14650v1}, and Mip-Splatting \cite{2311.16493v1}.
We initialized the gaussian environment map by sampling 100,000 points from a Fibonacci sphere. 
We set $\lambda_{sky}$ as 10 to remove cloud artifacts in the beginning 7,000 steps.

\subsection{Effectiveness of Gaussian Surface Reconstruction(GauS)}
For our gaussian surface reconstruction, we selected the SOTA method NeuS~\cite{neus}, SuGaR~\cite{sugar}, GaMeS~\cite{games}, and mesh extraction used in LGM~\cite{lgm} for comparison.
We utilized the camera poses of training images to render depth maps from GS and removed the pixels whose depth values are larger than 10.
To adjust the parameters of VDBFusion~\cite{vdbfusion} according to the size of input GS, we first calculate a bounding sphere and denote its radius as $r$. 
We set voxel size as $r/256$, SDF truncation value as $r/64$, and space carving as False in VDBFusion~\cite{vdbfusion}.
It is noted that if the camera poses of training images are unavailable, we render depth maps corresponding to the camera trajectory which circles the center of the scene.

We evaluates the visual quality of meshes generated by Gaussian Surface Reconstruction (GauS) compared to three other popular methods: SuGaR, LGM, and GaMes. Here, we assess how well each approach captures the intricate details and overall shape of the target surface.

Our observations suggest that our GauS achieves a good balance between quality and efficiency. While SuGaR could produce detialed surfaces, it suffered from large episollod , GauS maintains a sharp representation without a significant increase in processing time. Conversely, LGM may prioritize coarse shape with limited training time . GaMes often excels at capturing the local shape, but coarse structures might be underrepresented. GauS strikes a middle ground, offering visually compelling meshes that are faithful to the original surface while remaining free from excessive noise or over-smoothing. This balance between quality and efficiency makes GauS a compelling choice for gaussian surface reconstruction tasks.

\begin{figure}[h]
    \centering
    \includegraphics[width=\linewidth]{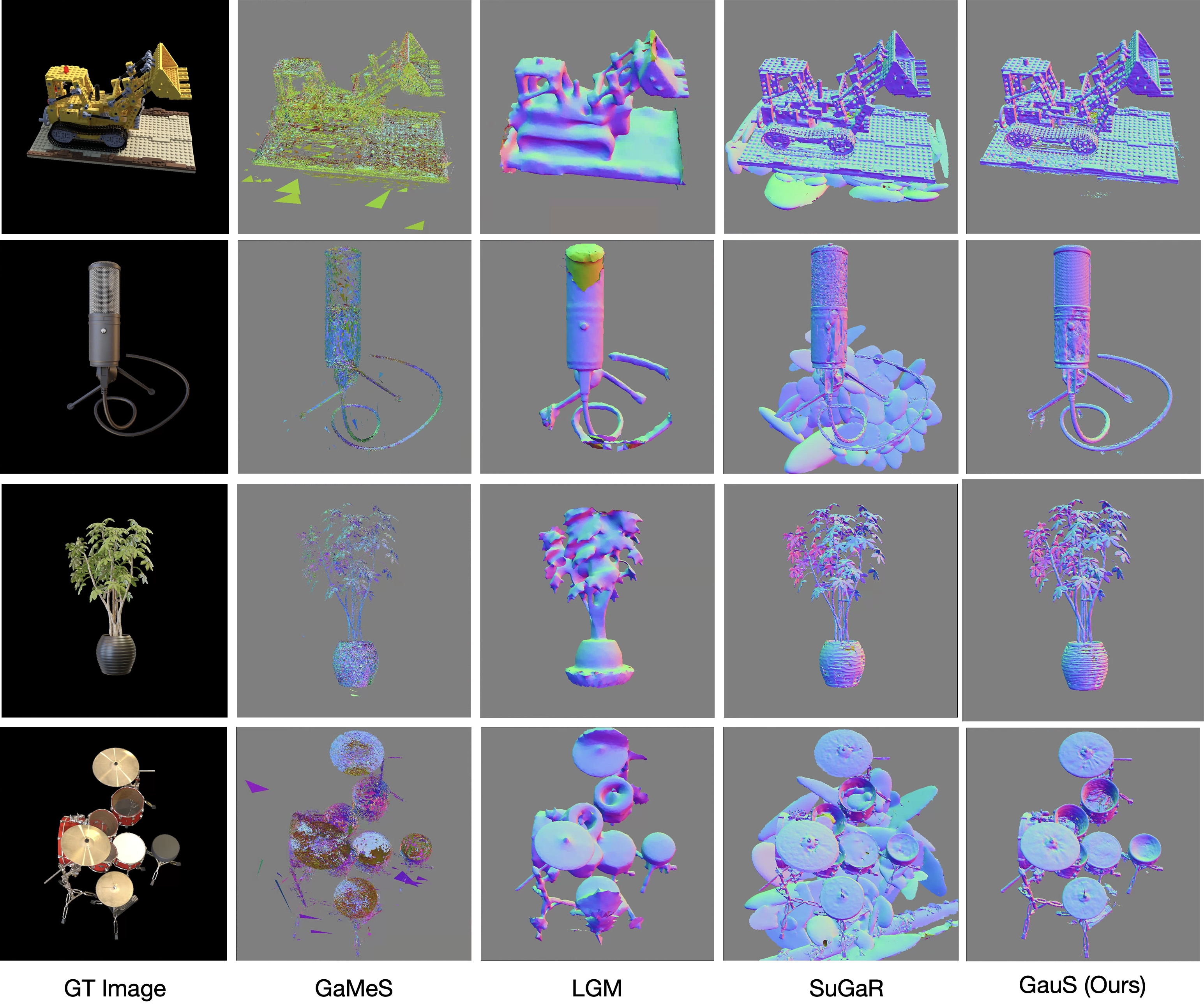}
    \caption{Qualitative comparisons of Surface Reconstruciton on Blender Dataset}
    \label{fig:skydome}
\end{figure}

Furthermore, we compared GauS with Sugar specifically on phone-captured data from the BlendedMVS dataset. Here, GauS demonstrated a superior ability to preserve both the fine details and the coarse structure of the object. While Sugar produced very fine-grained surfaces, it often lost some of the larger features and introduced additional artifacts. This highlights GauS's advantage in maintaining a balanced representation, particularly for data with varying levels of detail.

\begin{figure}[h]
    \centering
    \includegraphics[width=\linewidth]{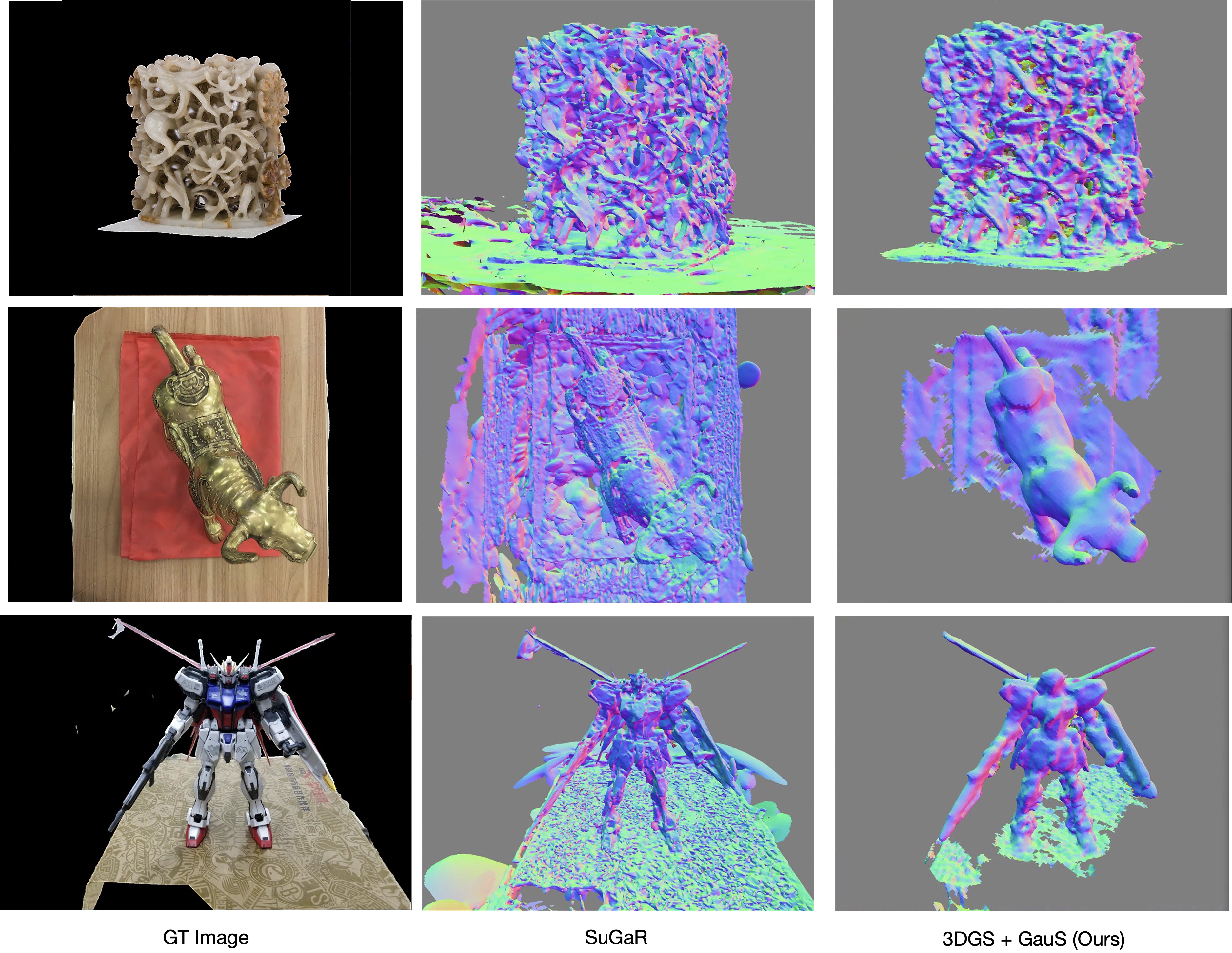}
    \caption{Qualitative comparisons between SuGaR and GauS on BlendedMVS Dataset}
    \label{fig:skydome}
\end{figure}

\begin{figure*}[h]
    \centering
    \includegraphics[width=\linewidth]{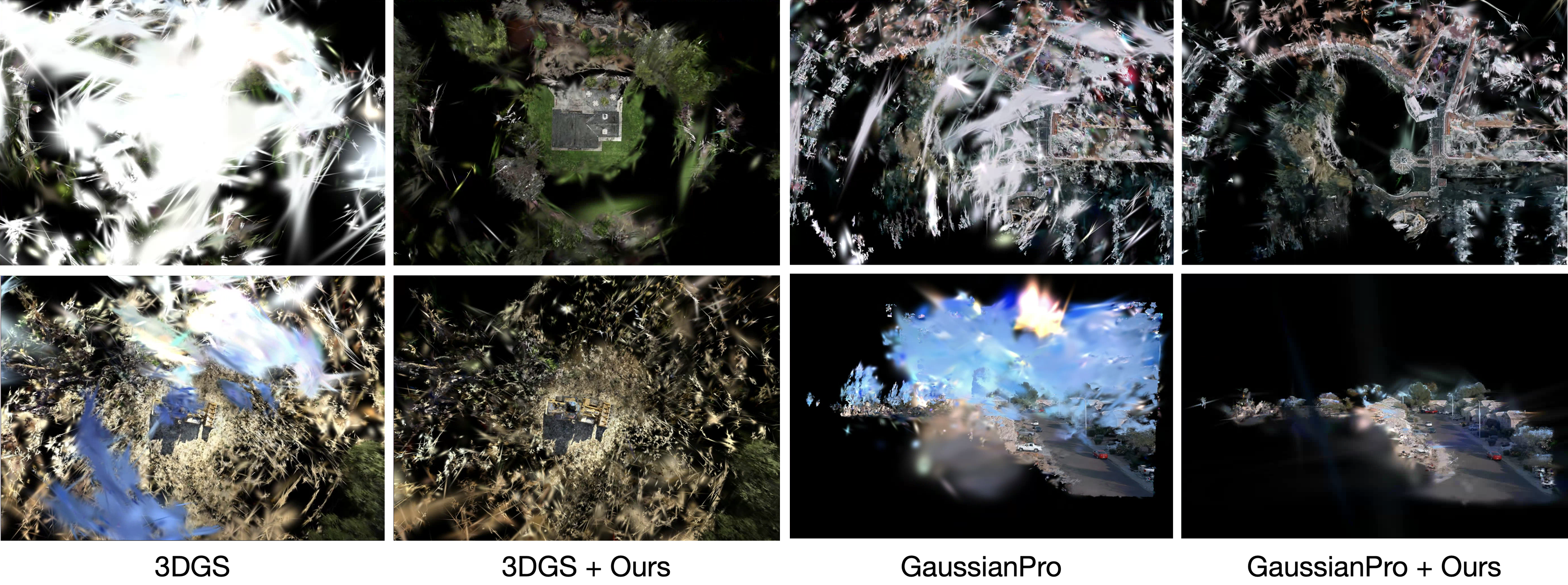}
    \caption{Qualitative comparisons of Gaussians Sky modeling on three types of outdoor scenes(unbounded, aerial, street). Ours refer to Gaussian Sky Modelling.    Our method effectively separates the sky from the foreground, reducing cloud artifacts and producing clean and realistic foreground representations.}
    \label{fig:skydome}
\end{figure*}

\begin{figure}[h]
    \centering
    \includegraphics[width=\linewidth]{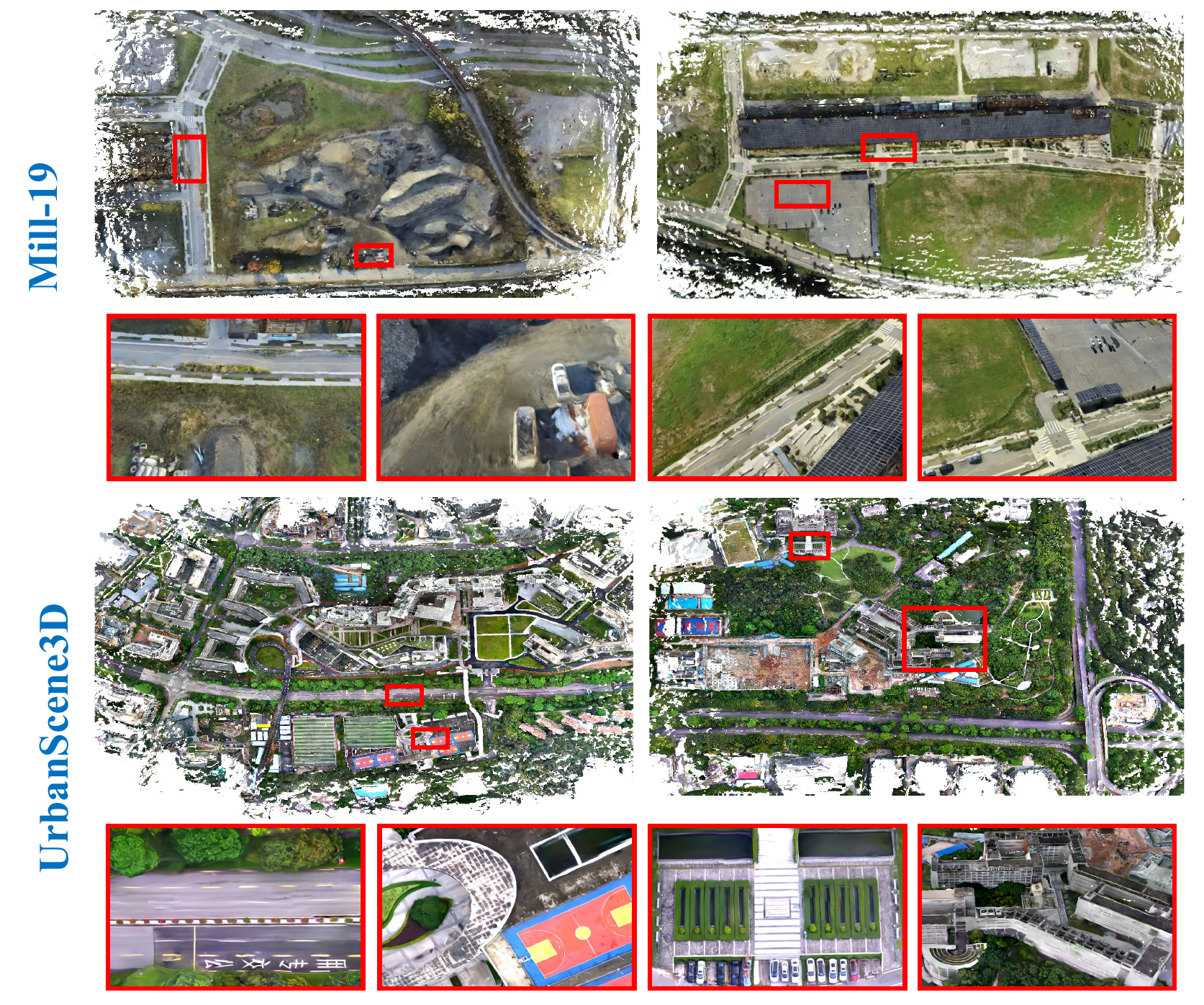}
    \caption{Qualitative results on the mesh extraction of large-scale scenes on Mill-19\cite{Turki_2022_CVPR} and UrbanScene3D\cite{UrbanScene3D} dataset.}
    \label{fig:large-scale}
\end{figure}

Finally, we extended GauS for application to large scenes, a challenging task for traditional NeRF and NeuS methods. NeRFs typically struggle with reconstructing large scenes due to their inherent limitations in handling global coherence. NeuSs, on the other hand, can handle large scenes but often require significant computational resources and may not always achieve the desired level of detail. By design, GauS offers a balance between these two approaches, making it suitable for tackling complex reconstruction problems beyond smaller objects. This opens up new possibilities for applications in areas like architectural modeling, autonomous vehicle navigation, and virtual reality scene creation, where reconstructing large and intricate environments is crucial. In Fig \ref{fig:large-scale}, we show the visualization of models extracted by GauS from the 3DGS model trained with appearance embedding\cite{2402.17427v1} and our sky modeling on Mill-19 and UrbanScene3D Datasets.

\subsection{Effectiveness of Gaussian Sky Modeling}
 We evaluate the effectiveness of our Gaussians Sky modeling approach by comparing it with two state-of-the-art methods, 3DGS \cite{3dgs} and GaussianPro \cite{2402.14650v1}, on three typical outdoor scenes. Fig. \ref{fig:skydome} shows representative examples from these scenes. Our method achieves a noticeable improvement in reducing noise and artifacts in the sky regions compared to the baseline methods. We further show quantitative comparisons on the Tanks and Temples dataset\cite{} in Table \ref{tab:tnt}.

In the unbounded scenes (left in Fig. \ref{fig:skydome}), 3DGS \cite{3dgs} suffers from prominent cloud-like artifacts in the sky, which can be distracting and detract from the overall realism of the synthesized view. In contrast, our approach effectively separates the sky from the foreground elements, producing a clean and artifact-free gaussian point cloud. 

In the aerial scene (top right in Fig. \ref{fig:skydome}), although GaussianPro \cite{2402.14650v1} also integrates sky masks to reduce floater accumulation in the sky region, it did not explicitly regularize the sky region and propose a proper model for the sky. This results in a partially cloudy reconstruction.

In the street scene (bottom right in Fig. \ref{fig:skydome}), our method generates a coherent street reconstruction, while the baseline methods exhibit visible artifacts and inconsistencies in the sky region.

\begin{table}[htbp]
\centering
\scalebox{0.8}{
\begin{tabular}{l|c|c|c|c|c}
& Barn & Caterpillar & Ignatius & Train & Truck \\
\hline
3DGS & 28.10 & 23.56 & 22.05 & 21.10 & 25.19 \\
GaussianPro & 27.66 & 22.25 & 20.79 & 20.72 & 23.58 \\
Scaffold-GS & 28.33 & 24.00 & 22.78 & 22.20 & 25.57\\
Mip-splatting & 28.30 & 23.64 & 22.11 & 22.15 & 25.25 \\
\hline
3DGS+Ours & 28.12 & 23.37 & 21.93 & 22.04 & 24.85 \\
\end{tabular}
}
\caption{Quantitative assessment on the Tanks and Temples dataset (PSNR). }
\label{tab:tnt}
\end{table}

These examples highlight the effectiveness of our Gaussians Sky modeling technique in improving the overall quality and realism of novel view synthesis for outdoor scenes. By explicitly modeling the sky as a separate component and incorporating semantic priors, our approach successfully mitigates common issues such as cloud artifacts, noise, and color inaccuracies, resulting in more natural and visually compelling renderings.



\section{Conclusion}
\label{sec:Conclusion}
In summary, GauStudio is a flexible and modular framework tailored for 3D Gaussian Splatting techniques. It allows customizable integration of different components for foreground modeling, background representations, and other modules to construct specialized pipelines for diverse 3D scene modeling tasks like reconstruction, editing, and simulation. The framework's key strength is its composability, enabling rapid innovation. GauStudio is complemented by GauS, an efficient module to extract textured meshes from optimized Gaussians across different methods. Overall, this comprehensive yet customizable platform aims to drive advancements in 3D scene modeling through its modular design and versatility.

\clearpage
{\small
\bibliographystyle{ieee_fullname}
\bibliography{ref}
}

\end{document}